# Towards Incremental Learning in Large Language Models: A Critical Review

Mladjan Jovanovic, Peter Voss

*Abstract* [*]—**Incremental learning is the ability of systems to acquire knowledge over time, enabling their adaptation and generalization to novel tasks. It is a critical ability for intelligent, real-world systems, especially when data changes frequently or is limited. This review provides a comprehensive analysis of incremental learning in Large Language Models. It synthesizes the state-of-the-art incremental learning paradigms, including continual learning, meta-learning, parameter-efficient learning, and mixture-of-experts learning. We demonstrate their utility for incremental learning by describing specific achievements from these related topics and their critical factors. An important finding is that many of these approaches do not update the core model, and none of them update incrementally in real-time. The paper highlights current problems and challenges for future research in the field. By consolidating the latest relevant research developments, this review offers a comprehensive understanding of incremental learning and its implications for designing and developing LLM-based learning systems.**

## I. Introduction

### A. Contribution and Structure

Incremental learning (IL) is the ability of a learning system to make decisions based on continuously incoming inputs rather than relying only on previously processed data. This allows the system to adapt and respond to changing conditions and user behavior immediately rather than waiting for periodic retraining.

Given the paradigm's complexity, the existing LLM solutions aim to enable IL but still haven't achieved it in practice. Therefore, this review takes a broader perspective on incremental LLM updates, considering them batch-incremental periodic model updates. It also differentiates between core model updates and external system updates.

Thus, the critical review primarily discusses IL approaches, methods, and techniques in Large Language Models (LLMs) that acquire new abilities to solve unseen tasks while preserving existing ones.

Distinct from typical systematized reviews, the paper highlights the following key factors:

1) *Analysis of topics related to IL in LLMs.* The paper covers fundamental aspects of continual learning, meta-learning, parameter-efficient learning, and mixture-of-experts learning, including learning from scarce, changing, or non-existent task information, learning multimodal tasks, and learning task sequences. By exploring the IL in related learning paradigms, we highlight the challenges and opportunities in these areas.
2) *Comprehensible and compact exposition.* This paper recognizes the importance and complexity of IL and describes the concepts, methods, and techniques clearly and concisely. It aims to reach a broad audience, including engineers and researchers. Through brief descriptions, explanations, and references to original works, the paper helps readers grasp fundamental IL principles and their practical implications.
3) *Synthesis of essential information*. IL is a fast-growing field, with information scattered across various fields and sources. The paper consolidates the relevant information about IL, offering a comprehensive overview to interested readers. Synthesis of the latest practical developments aims to guide researchers and engineers looking for a deeper understanding of IL and its LLM-based applications.

By highlighting these contributions, the paper complements existing related reviews and offers a unique perspective into the current situation and outlooks for IL in LLMs. The paper consulted relevant sources, including Scopus, ACM Digital Library, and IEEE Xplore Digital Library. The inclusion criteria for each paper analyzed is the practical verification of the proposed approach describing the solution, learning process, training data, and experiment results.

This paper provides critical insights into existing IL approaches within the LLM design space. To do so, we first define the key LLM concepts and notations used throughout the paper in the current section. Section 2 delves into the current state of IL approaches from a unified perspective by categorizing them into three related topics: continual learning methods, meta-learning methods, parameter-efficient tuning methods, and a mixture of expert methods. The paper describes the representative examples from different approaches. Section 3 extracts the main findings with implications concerning the application of IL to real-world problems with an overview of the challenges ahead. The concluding section summarizes these challenges and research opportunities.

### B. Context and Motivation

LLMs are based on a transformer neural-network architecture [1] and produce text in natural language. Transformers introduced the *attention mechanism* that tracks the relations between words across lengthy text sequences in forward and reverse directions. It pays attention to (co-) appearances of words as chunks of information in terms of probabilities, not the meaning of those words. Thus, it cannot *understand* since there is no reference to the "real world" or "conceptual world" to ground as explicit knowledge. Instead, they approximate complex probability distributions on a large number of variables (the context of a word in text) and predict next words by drawing from these probability distributions. That said, it represents domain-specific knowledge as a language structure relying on context-dependent long-range associations (aka parametric knowledge).

Over time, the architecture expanded for generating other media as embedding vectors have become numerical representations of text, audio, image and video (i.e., encoders for different data modalities) [2, 3].

LLMs can produce plausible human-like content in various domains. Models with these generative capabilities are called Foundation Models (FMs). They learn about existing data artifacts and use learned patterns to generate new data instances, including text, images, music, design, and motion. Massive models process vast amounts of unlabeled data, prioritizing general-purpose applications. These models are then customized to create tools and infrastructure for specific domains. This way, FMs can be adapted across a variety of scenarios. Table 1 lists fundamental LLM architectures.

TABLE 1. FUNDAMENTAL LLM ARCHITECTURES.

| Type | Application | Learning | Models |
|---|---|---|---|
| Encoder (text-features) | Natural language understanding (NLU) | Masked language modeling | BERT, DeBERTa, ALBERT |
| Decoder (features-text) | Natural language generation (NLG) | Language modeling | GPT4, ChatGPT, Gemini, LLaMA, Falcon, Bard |
| Encoder-decoder (text-features-text) | Sequence-to-sequence tasks | Sequence-to-sequence modeling | GLM, T5, BART |

The encoder transforms input text into a fixed-size set of features as words' positional encodings (embedding vectors) for NLU tasks, including named entity recognition and sentiment classification. The decoder accepts features to produce output text, such as dialog, question answering, and text summarization. Combined architecture encodes the input text into vectorial representation and decodes embeddings to generate output, such as machine translation.

The transformer-decoder architecture is shown in Figure 1. It consists of the embedding component, a decoder stack, and a head component with linear and softmax layers. The embedding component transforms NL text into numerical vectors (or tokens) chunks. The tokens are forwarded to the decoders comprising Multi-head Self-Attention (MSA) and Feedforward Network (FFN) modules. The MSA clusters each token by an attention map obtained from two linear mappings of the input tokens (as their dot product). An FFN then processes the tokens. LLMs stand out from other Deep Neural Network (DNN) architectures in two ways. First, they exhibit the autoregressive pattern, performing the generation task in iterations. Next, they incorporate the attention mechanism, which quadratically scales computational complexity with the input length. At the same time, LLM's inherent computations occur in the attention blocks within each decoder layer. LLMs iteratively generate tokens (words) based on the previously generated sequence. The process repeats until the model processes and outputs the entire input sequence.

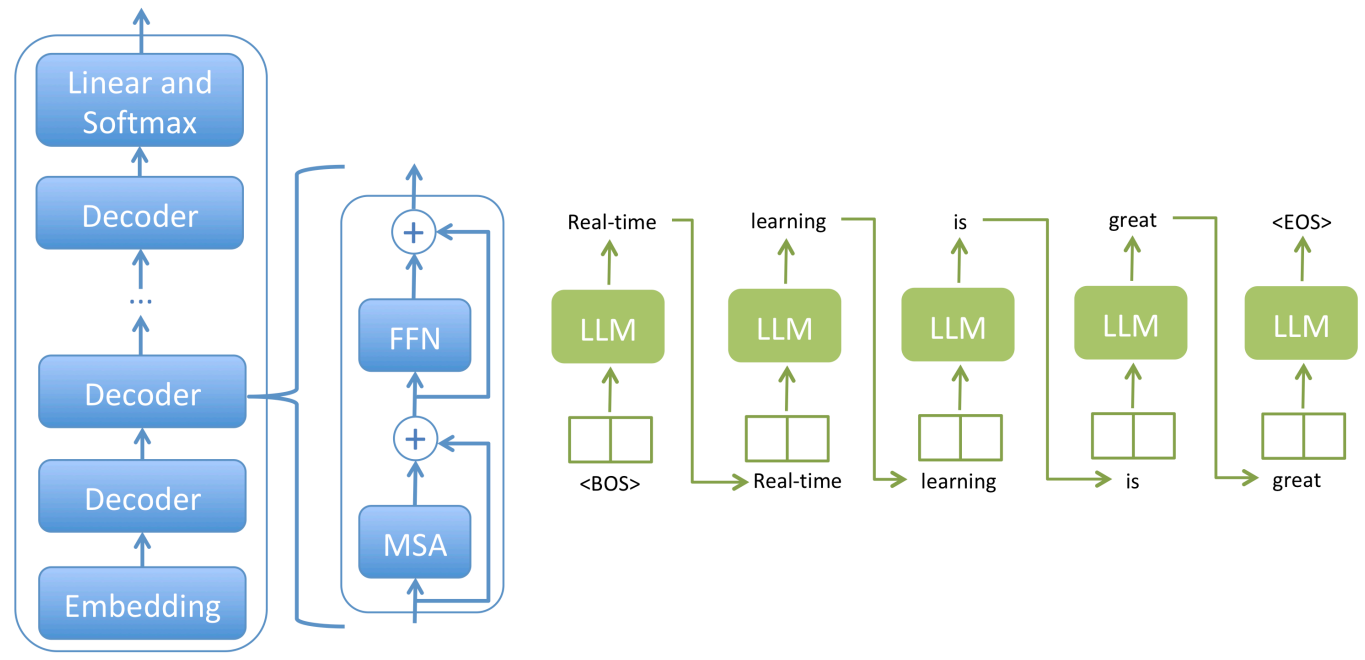

**Fig. 1.** The transformer-decoder architecture (left), and auto-regressive behavior (right). Adapted from [1].

Typical LLM learning strategies are as follows [3, 4]:

1) *Pretraining-finetuning* includes unsupervised training with a massive amount of data for general NLP tasks (also known as upstream tasks), followed by supervised training on a smaller amount of domain-specific data for specific tasks (aka downstream tasks).
2) *Instruction tuning* involves obtaining a new, instruction-financed model capable of following instructions. The model learns new tasks via natural language instructions. It involves actual training of the model by changing its weights. *Soft prompt* tuning is a case in which the model is frozen, and only the soft prompts (learnable tensors obtained through backpropagation) added to the input embedding are learned [5]. Unlike finetuning, which involves billions of parameters to train, in soft prompting, there are typically several thousand parameters to train. Similarly, *prefix-tuning* optimizes a small continuous task-specific vector (aka prefix) where subsequent tokens attend to the prefix [6]. It obtains state-of-the-art performance compared to full-data training and outperforms finetuning in low-data training by learning only 0.1% of the parameters [6].
3) *Prompting* refers to conceiving a prompt (as user input) that will yield the optimal response in a given situation. No training is involved (i.e., the LLM's weights are not changed due to no gradient updates). A prompt is a flexible task specification containing NL descriptions of intentions with examples and expected responses [7]. Various templates have been proposed as prompting techniques, including zero-shot (instructions with task-specific formatting but without task-specific examples), few-shot (instructions with a few task-specific demonstrations to guide the model's outputs), chain-of-thought (stepwise instructions with logical context to simulate reasoning), and tree-of-thought (guidance to select among multiple solutions) [7, 8, 9, 10]. *In-context learning* (ICL) is sometimes used to refer to few-shot prompting as the model is guided during inference with prompts containing task examples [7].

*Reinforcement Learning from Human Feedback* (RLHF) represents an alternative in which the model is refined based on human feedback or corrections [11]. Users express their preferences among different responses provided by the model.

Subsequently, a reward mechanism is established. The model enters a reinforcement learning loop to maximize the rewards by continuously refining its responses to better match the user's preferences. Ranking the model's responses from human evaluators offers a superior objective function compared to those utilized in pre-training. A recent study [12] highlighted practical challenges in using RLHF to customize LLM as misaligned humans (evaluators holding different goals and values), quality control with difficult oversight (humans making mistakes), representative data (biased), and challenges with reward model (reward hacking and evaluation). This often requires retraining these models, which is impractical regarding computation, time, and privacy.

*Retrieval-augmented generation* (RAG) [13] is a framework for increasing the quality of LLMs' outputs by grounding the models on external knowledge retrieval components to supplement the LLM's internal representation of information. Implementing RAG verifies LLM-based outputs against reliable facts and sources, ensuring users can check its claims for accuracy and trust. RAG helps reduce hallucinations, outdated knowledge, and non-transparent, untraceable content generation process [14, 15].

The resources required by LLMs (compute, data, environment impact) are enormous as they are based on statistical pattern matching on large quantities of human intelligence outputs. At the same time, LLMs are still out of proportion to achieving human-like intelligence, such as planning, reasoning, and self-verification [16, 17, 18].

IL of knowledge emerging in the real world is essential to achieve human-like intelligence [19]. However, standard neural network (NN) topologies are hindered by catastrophic forgetting (CF), a limitation that prevents them from learning a sequence of tasks. This issue must be addressed for an NN to successfully adapt to lifelong or continuous learning, a fundamental requirement for AI. Two common goals of continuous learning are to learn new tasks from known classes (online learning) and learn unknown classes (class-incremental learning) [20].
Rehearsing techniques can avoid CF [21]. This implies that when new data is introduced, the NN gets retrained on some previously learned data. However, previously learned knowledge may not be available for such retraining in general. In addition, CF may occur when compensating for concept drift [21]. Deep Learning (DL) assumes that a training set is a representative sample of the underlying unknown distribution. If the distribution space shifts, new modeling should expand this shift. If the initial space is merely shifted instead of adjusting the size of the distribution space, then CF will occur.

The current situation inspired us to look deeper into existing work on topics related to IL in LLMs, including continual learning, meta-learning, parameter-efficient learning, and mixture-of-experts learning, and the challenges posed by the extensive parameterization of LLMs and continual model adjustments to address deficiencies or undesirable behaviors.

## II. Related Work

### *A. Continual Learning*

Continual learning (CL) learns new, emerging tasks efficiently while mitigating CF in LLMs.

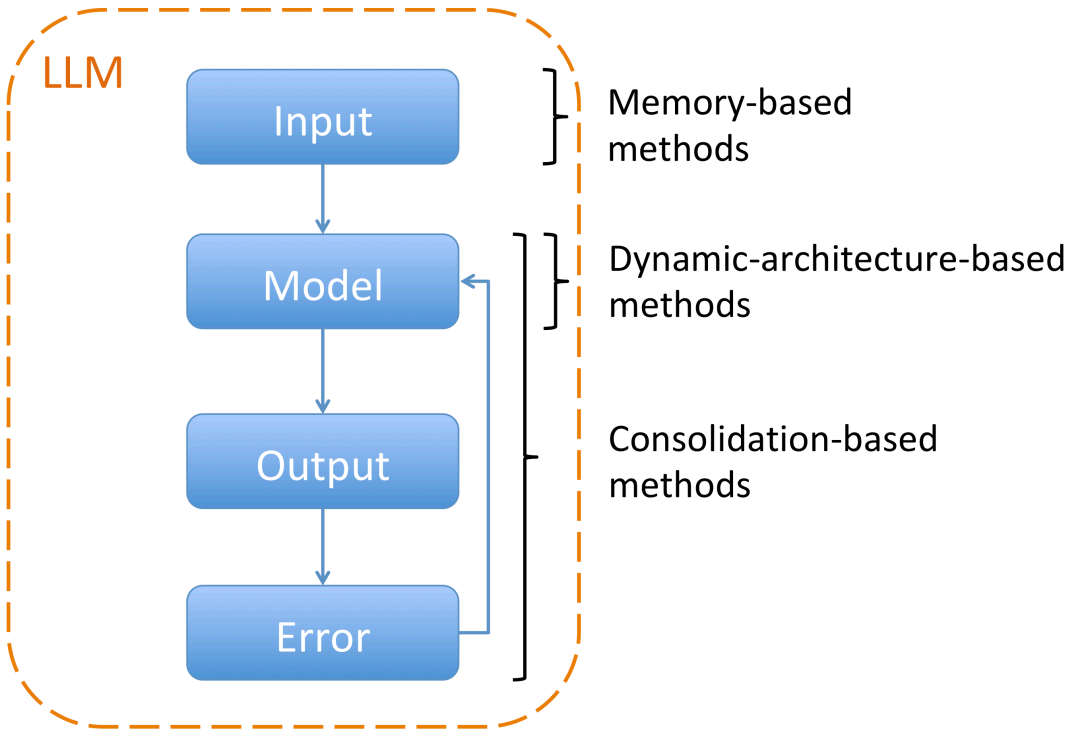

**Fig. 2.** Approaches to Continual Learning.

Figure 2 illustrates approaches to CL. The existing work on CL can be classified as follows [22, 23, 24, 25]:

1) *Consolidation-based* methods, such as *regularization* [26] or *distillation* [27], protect important parameters from significant shifts. They do this by aligning the current output space with previous ones (distillation loss) or restricting the model parameters by estimating loss or averaging weights (regularization loss). *Weight regularization* concerns parameters, whereas *function regularization* influences behaviors of the existing model.
   *Weight regularization* selectively regularizes variation of NN's parameters. It typically penalizes each parameter's variation based on its importance in performing the previous tasks [28]. *Function regularization* is a robust approach targeting the model's in-between results or the prediction. It typically implements Knowledge Distillation (KD) [29] by learning a small student model from a large teacher model previously learned to mitigate CF. The student model can learn from new training examples, replay previous examples, and use available unlabeled data or data generated from previously learned tasks.
   Aside from regularization, some methods focus on *error optimization*, such as calculating gradient propagation on the old tasks' input data [30] (e.g., rectifying the current gradient directions through parameter updates), meta-learning of arriving tasks [31] (e.g., obtaining an inductive bias for new tasks internally, rather than specifying it explicitly), and investigating loss spaces [32] (e.g., discovering a local minimum by adjusting the training variables, including learning rate decay, dropout rate, and batch size).
   Self-supervised learning (generating implicit labels from unstructured data) and upstream pretraining with downstream finetuning (pre-trained representations fixed when learning downstream tasks) also belong to this class as they create and employ internal representations in LLMs. However, they require large amounts of data.
   *Task-agnostic distillation* (TAD) iteratively prunes the student's model without relying on task-specific exemplars until it reaches the target size, thereby reducing the performance gap between the teacher and student models across a wider set of tasks [33]. It is also used within RL as a method for distilling a policy model in a self-supervised manner, without external goals [34]. Some approaches combine TAD variations to distill a compact student, such as extracting teacher parameters randomly or based on a relevance metric [35].

While these methods may only partially utilize historical data, their ability to safeguard crucial parameters is a significant advantage.

2) *Dynamic-architecture-based* methods [36] train a model by increasing its size (number of parameters) with the task number (also known as *parameter expansion*). These methods introduce task-specific parameters through a model adaptation by adding task-specific parameters (*parameter allocation*), creating task-specific modules (*modular network*), and differentiating shared and task-specific model components (*model decomposition*).

   *Parameter allocation* dedicates a parameter set in the model to each task. The expansion of such models should be balanced with the increase in tasks [37]. *Model decomposition* separates a model into shared and task-specific elements. For instance, the parameters can be decomposed by additive decomposition, singular value decomposition, and low-rank factorization [38]. *A modular network* includes modules or sub-networks learning incremental tasks simultaneously without sharing task-specific components. Mixture-of-Experts (MOEs) architectures are examples of this method. We describe them separately in Section 2.4.

   *Visual Prompt Tuning* (VPT) [39] combines previous techniques by utilizing lightweight trainable modules to adjust the pre-trained model by prepending learnable parameters (aka prompts) to the base features. The model concatenates parameters and features for the vision transformer blocks. Encoding task-specific information into the prompts while keeping pre-trained weights unchanged minimizes the cross-entropy loss.

   *Growing neural networks-based approaches* propose a scalable network of NN architectures in which specific nodes encode particular tasks, such as multitask learning in robots [40]. Alternatively, self-organizing incremental NNs can learn classification tasks while mitigating CF [41].

   The previous methods have a major drawback - the linear growth of training costs [22, 25] as the model size grows with new tasks linearly. This cost increase can be a deterrent despite their ability to overcome CF.

3) *Memory-based* methods [42] keep additional memory of examples of previous tasks, which are used (replayed) when learning new tasks. While they retain some information from previous tasks, they face challenges such as overfitting and low scalability due to frequent retraining of the model (aka *replay-based* or *rehearsal-based*).

   These methods typically approximate and reconstruct old data distributions. *Experience replay* records a small amount of old training examples [43]. *Generative replay* employs a generative model to produce synthetic examples for training [44]. *Feature replay* restores the distribution of old features [45] through saving prototypes (i.e., feature distillation between the old and new model or representation shift to update the preserved old features), statistical information (e.g., mean and covariance), or training a generative model.

   *Resource-efficient replay approaches* aim to merge similar inputs and increase distances between samples to enlarge sample variability and reduce memory [46]. Similarly, in streaming learning, a reduced number of prototypical examples that capture most of the variance within different classes can be stored [47].

   However, selecting exemplars from pretraining datasets may not be possible or can be difficult, and downstream data may not always be suitable for preserving the pretraining knowledge.

Table 2 gives an overview of CL methods, techniques, target model elements, and trade-offs.

TABLE 2. CL METHODS COMPARATIVE SUMMARY.

| Method | Technique | Element | Trade-offs |
|---|---|---|---|
| Consolidation-based | Distillation<br>Regularization<br>Optimization | Parameter subsets<br>Intermediate values and representations | Updates/extracts relevant parameters<br>Increased computation |
| Dynamic-architecture-based | Parameter/network expansion<br>Model decomposition | Task-specific parameters, modules and components | Updates portion of parameters<br>New tasks increase model size |
| Memory-based | Historical and derived task examples | Parameters<br>Components<br>Layers | Retains model size<br>Obtaining suitable training samples |

Some definitions of IL refer to continual learning techniques in NN-based models [23] as task-incremental, domain-incremental, and class-incremental learning, depending on whether task identity is specified at test time or is inferred by the model. Task-incremental learning (TIL) learns to solve several distinct tasks. Domain-incremental learning (DIL) learns to solve the same problem in different contexts (input distribution changes). Thus, the context refers to the underlying distribution from which observations are sampled. Class-incremental learning (CIL) learns to discriminate between incrementally observed classes. The main challenges in domain-incremental and class-incremental learning are preventing forgetting and correctly identifying tasks, respectively [48].

IL in LLMs can also be classified as offline IL and online in-context learning [49], considering training data sources.

Offline IL, while allowing for specific adaptation with new annotated data sets (e.g., supervised fine-tuning and RLHF [11]), has its limitations. It requires frequent model updates for evolving tasks and scenarios, and static post-updates are inflexible to adapt to real-time changes.

Online in-context learning assumes an LLM interacts with an external AI model or agent or with an offline or online knowledge source (e.g., RAG [13]). Such methods depend on external knowledge and do not retain persistent learned information, which may lead to a loss of used knowledge.

Existing reviews examine CL as LLM adaptations across time, domains, and capabilities by focusing on the kind of information that is updated and learning strategies at their specific training stages (i.e., pre-training and fine-tuning) [22, 50, 51]. Specifically, the analysis from [51] classifies CL as:

- *Continual Pre-training* (CPT) expands the model's general NL understanding and generation capabilities [52]. It conducts self-supervised training to enrich the model's general NLP knowledge and general knowledge of new domains, such as learning new natural or programming languages, factual knowledge (e.g., most recent information), and domain knowledge (e.g., e-commerce).
- *Continual Instruction Tuning* (CIT) improves the model's response to specific user inputs [53]. It conducts supervised instruction-following finetuning to learn domain-specific tasks. For instance, learning to solve new tasks (e.g., code generation), tasks in novel domains (e.g., medicine), or to use new tools to solve problems (e.g., search engines).
- *Continual Alignment* (CA) ensures the model's outputs adhere to human values, preferences, and ethical and societal norms [54]. For example, learning to align to ethical and social norms (e.g., culture-specific norms) or users' preferences (e.g., emerging values on new users).

While it does not provide an in-depth analysis of model workings and training data, our review aggregates their principles and criticalities according to the proposed classification. Moreover, it outlines some typical cases, reveals related effects, and highlights practical limitations for a broader audience as follows. While some examples combine different methods, we classify them based on the primary method employed.

*Consolidation-based examples*

Prototype and Low-Rank Adaptation (PLoRA) [55] starts from a pre-trained backbone and finetunes a small number of parameters to learn new classes in Federated Class-Incremental Learning (FCIL). FCIL is a method to continuously learn new classes while addressing the challenge of forgetting old classes [56]. It is based on a distributed ML paradigm called Federated Learning (FL), where multiple data owners (or local models) collaborate to train a shared or server model. Existing methods [56, 57] propose different loss functions on the client and server sides to alleviate local and global CF but result in significant memory and communication costs. PLoRA introduces a shared prototype module for the global server, aggregating (re-weighting) the local prototypes from clients. The local prototype vectors are derived from prototypes of corresponding class features. Then LoRA [58] finetunes only trainable parameters and sends them to the global server. In particular, each client uses local data to train LoRA parameters to obtain prototypes of each class. Only the trainable parameters and the average class feature are shared with the server. The server re-weights the local prototypes it received to form the global prototype. PLoRA utilizes ViT-B/16-IN21K [59] as the backbone PTM for server and client models. While PLoRA outperforms state-of-the-art approaches, it has high training costs for data, memory, and computational resources for server and client models.

Few-Shot Class-Incremental Learning (FSCIL) [60] trains a base model with sufficient data. Later, it utilizes the knowledge of base classes to learn of new classes with limited data while preserving the understanding of old ones. FSCIL usually fine-tunes the entire model, being susceptible to overfitting and with reduced performance in learning new classes [60]. The Attention-aware Self-adaptive Prompt (ASP) framework [61] adds specific prompts between attention blocks as attention-aware task-invariant prompts (TIP) and self-adaptive task-specific prompts (TSP). The TIP contains task information common for all classes (old and new). The prompts are constructed by encoding input images strongly correlated with semantically relevant information and weakly correlated with irrelevant image information. The TSP is constructed for a particular input image, combining the average and image prompt features. This way, it fine-tunes only the task-specific parameters. ASP uses ViT-B/16-1K [59] as the base model, pre-trained on the ImageNet1K dataset, which spans 1000 classes and 1,281,167 training images, 50,000 validation images, and 100,000 test images. The ASP outperforms state-of-the-art FSCIL methods in learning new classes. However, training costs are high, and learning new classes can drop the performance of previous ones.

The INCremental Prompting (INCPrompt) [62] utilizes flexible task-aware prompts to effectively represent task-specific information and mitigate the CF of old tasks. It implements an attention-based transformation of the token sequence whose results are passed to a group of FF layers with a non-linear activation function (named the prompter) to generate the prompt as key and value components. Experimental evaluations on Split CIFAR-100 (60,000 images in 100 classes) and Split ImageNet-R [63] (30,000 images in 200 classes) CL benchmarks showed improved performance over existing methods. However, its scalability concerning task number and data complexity and generalizability to other CL scenarios with different data distributions remain unclear.

Incremental Vision-Language Object Detection (IVLOD) [64] adapts pre-trained Vision-Language Object Detection Models (VLODMs) to different domains while preserving their general capabilities. The method employs Zero-interference Reparameterizable Adaptation (ZiRa) with Zero-interference Loss (ZiL) and reparameterization techniques for IVLOD without incurring computation and memory costs. In particular, it retains the original model parameters and adds a parallel branch component (Reparameterizable Dual Branch - RDB) for tuning on downstream tasks without overwriting existing knowledge. Based on the RDB's dual-branch structure, the ZiL component reduces the interference of new knowledge on learned knowledge by penalizing the RDB's output. The RDB learns new tasks while protecting pretraining and downstream task knowledge. The IVLOD uses a Swin Transformer [65] as a backbone VLM. The evaluation demonstrates that ZiRa preserves general capabilities while accurately learning new tasks, outperforming state-of-the-art object detection methods [66, 67]. However, the enhanced performance comes with an increased model size.

The study on *Knowledge Editing* (KE) for LLMs presents practical and efficient methods for modifying and evaluating LLMs post-training to learn new tasks or correct outdated information [68]. These methods, such as knowledge insertion, modification, and erasure, are designed to allow on-the-fly model modifications while maintaining overall performance across various inputs.

In analogy to the human learning types, KE methods are categorized as:

1) *Recognition* - exposing the model to new knowledge within a relevant context.
2) *Association* - forming connections between the model's new knowledge and previous knowledge in the model.
3) *Mastery* - the model acquires and utilizes the knowledge through its parameters.

However, the above methods focus on LLMs' components - memory, retriever, layers, and parameters which reflect correlations rather than explicit knowledge. In particular, recognition as resorting to external knowledge is implemented by expanding LLMs' layers (e.g., FFNs) and parameters (e.g., LoRA) [69]. The association is realized via user interactions with the LLM without retraining as additional memory of instruction examples [70] (i.e., various prompting techniques). Mastery manifests through cost-efficient alternatives to fine-tuning. For instance, meta-learning methods edit the model by learning the change in the model instead of updating the weights directly. Model Editor Networks using Gradient Decomposition (MEND) [71] uses the rank-one decomposition to split the model into separate rank-one matrices to compute the change of the weights, reducing the number of parameters. Some techniques try to locate where the knowledge was stored and edit that part of the model [72] (e.g., FFN's area or a set of model weights). However, the side effects are unclear due to underlying LLM opaqueness [73].

The study in [74] introduces a post-training adjustment method to mitigate CF and improve CL in MLLMs. The Model Tailor retains the pre-trained parameters and replaces fewer finetuned parameters (up to 10%). Building on the tailor metaphor that selects patches for a garment, it identifies and adjusts a minimal parameter set as a sparse mask (or the model patch) from the finetuned model. It extracts the patch by inspecting parameter and loss changes to identify a critical subset of finetuned parameters essential for learning a target task. Moreover, it decorates the patch by compensating weights with an inverse Hessian matrix to enhance the performance on both current and previous tasks. The method is evaluated using InstructBLIP [75] and LLaVA-1.5 [76] MMLMs for VQA and image captioning tasks with available benchmark datasets. The results show improved CL while preserving pre-trained capabilities compared to traditional finetuning. However, it remains to be verified in more complex and open domains.

An approach to ingrain learning-in-realtime capabilities is an integration of LLMs with knowledge graphs (KGs) [77]. Neurosymbolic computing [78] combines neural networks' (NNs') pattern-recognition capabilities with KGs' reasoning abilities by either compressing knowledge structures into vectorized representations suitable for NNs [79] or linking neural patterns with symbolic knowledge by extracting pertinent pattern information [78]. Currently, the LLM-KG integration is mainly implemented in learning pipelines with separate LLM and KG components, losing some of the original semantics in the produced representations in both techniques [77, 80].

Continual Optimal Policy Regularization (COPR) is a method that aligns LLMs with human preferences in a CL setting [81]. RLHF-based techniques often require retraining when encountering new queries or feedback, as human preferences vary across domains or tasks. The method computes the distribution of optimal policy bypassing the partition function and then regularizes the policy based on optimal distribution to mitigate CF and improve preference learning. Although the method outperforms existing CL methods in the task and domain CL settings, it still needs RLHF reward and policy models to start from, which may reflect the preferences of a particular user group and induce a bias when optimizing the current policy.

While LLMs' attention has a quadratic dependence on context size, it is often sparse where a query is most relevant to a subset of the keys, such that many similarity scores converge to zero and their computation can be skipped. While predefined heuristics primarily drive existing sparsity solutions, intrinsic attention adaptation [82] learns the attention sparsity directly from the LLM itself as a gate that selectively activates a subset of relevant attention blocks. The technique is computationally efficient and improves output accuracy. Yet, the accuracy drops with context size.

*Dynamic-architecture-based examples*

The study described in [83] employed various techniques to prevent CF of the finetuned LLaMA2 model with 7 billion parameters, including freezing specific model layers, freezing specific model modules (e.g., self-attention and FF modules), and model adapters introducing additional trainable parameters (e.g., LoRA [58]) The evaluation results showed the repetition problem (the greedy approach of choosing the next token, where the highest probability token is always selected, causes the model to generate a repetitive token sequence) and a decline in reliability. In contrast, the model's knowledge remained mainly unaffected. However, the model was evaluated on only two distinct tasks.

Excessive LLM modifications can cause forgetting, while insufficient adjustments make inadequate fits for new classes. ExpAndable Subspace Ensemble (EASE) is an approach to Pretrained Model (PTM)-based CIL addressing the above issue [48]. The expanded model initializes and trains an adapter with trainable parameters encoding task-specific information per incoming task. The model extracts embeddings by aggregating historical features to synthesize prototypes of old classes later. In addition, all adapters extract the prototypes of the current dataset for new classes. Verification using ViT-B/16-IN21K [59] (a vision transformer encoder model with 86.4 million parameters pre-trained on ImageNet-21k containing 14 million images with 21,843 classes) as a backbone PTM shows state-of-the-art performance. Although adapters are small (0.3% of the base model), model size increases with additional adapters.

MiniGPT-4 [84] connects a pre-trained visual encoder and a pre-trained LLM. In particular, it combines a vision encoder (a pretrained ViT coupled with Q-Former [85]), Vicuna LLM [86], and a projection layer connecting the two models. The linear projection layer aligns the visual features with the textual ones. It does so in a two-step training method. First, it pretrains the vision encoder on many image-text pairs to learn vision-language skills. Then, it finetunes the model using a small number of high-quality image-text examples and a predefined prompt template to elicit more natural and reliable NL outputs. MiniGPT-4 showcases compositional vision-language abilities in a variety of tasks (e.g., generating detailed image descriptions, extracting facts from movie photos, and image captioning). However, the model inherits the backbone LLM's limitations and may hallucinate (e.g., identifying non-existent objects in the image), especially with longer captions.

The Generative Multi-modal Model (GMM) framework for CIL [87] generates detailed image descriptions by incorporating visual and textual information with a customized generative model. The model contains an encoder producing image and text embeddings image and text, which are further processed by an auto-aggressive decoder to generate image descriptions. Then, it uses a text encoder to produce textual features from the image descriptions and matches features to decide on a label that is highly similar to the prediction. The framework employs Bootstrapping Language-Image Pre-training with frozen unimodal models (BLIP) [88], a vision-language model pre-trained from existing pre-trained image encoders and LLMs with significantly fewer trainable parameters than existing methods. The model bridges modalities using a lightweight Querying Transformer learning vision-language representation with the image encoder and vision-to-language generation with the language model. GMM improves the state-of-the-art classification accuracy and CF. However, the main challenge is transferring the generated textual information into classifying distinct categories.

The study in [89] presents an Interactive Continual Learning (ICL) framework inspired by the Complementary Learning Systems theory [90]. ICL utilizes interactions among different models. Specifically, it explores the interactions between ViT (simulating System1) and MLLM (being System2). It adapts System1 parameters while keeping System2 parameters unchanged. System2 handles complex examples and interacts with System1. The Class-Knowledge-Task Multi-Head Attention (CKT-MHA) module enables continuous ViT finetuning using category features and from Class-Task collections. This way, System1 retains old parameters, avoiding ineffective updates. The von Mises-Fisher Outlier Detection and Interaction (vMF-ODI) mechanism estimates sample complexity so that System1 can select complex enough examples. The examples are used for System1 inference, whose predictions represent background knowledge for System2, thereby coordinating the two systems. System 1 uses ViT-B/16-IN21K [59] as a backbone, whereas System 2 employs MiniGPT-4 [84] as a base model. ICL was evaluated using CIFAR-10 (10 classes, with 50,000 training and 10,000 test color images per class), CIFAR-100 (100 classes, 500 training, and 100 testing images per class), and ImageNet-R (200 classes, 30,000 images). The evaluation showed CF resistance and superior TIL and CIL performance compared to existing methods. However, the limitations of the backbone models restrict the performance enhancement of System2.

Research from [91] pre-trained and evaluated the T0 model with 3 billion parameters [92] on 50 datasets related to 50 distinct textual QA, classification, and summarization tasks (100,000 examples for training per task). The model was extended to and verified through CL on 8 new language generation tasks (e.g., headline generation, text simplification, and haiku generation). On the one hand, the resulting model demonstrated acceptable performance in learning new tasks (e.g., concept drift) while maintaining nearly 100% of the initial performance. On the other hand, the study showed that CL emerges from the intensive pre-training stage.

Progressive Prompts [93] is a technique of learning a soft prompt per task, concatenating it with the learned prompts while keeping the base model unchanged. The goals are to keep the knowledge of previous tasks and allow the transfer of existing knowledge for new tasks. The model parameters consist of the fixed PLM parameters and fine-tuned prompt parameters. The study used an encoder-only BERT model and an encoder-decoder T5 model on text classification tasks. The proposed approach eliminates the need for data replay or storing many task-specific parameters. It shows superior performance on a standard text classification CL benchmark on longer CL sequences spanning 15 tasks. Nevertheless, the model size grows when the number of new tasks increases.

Visual instruction tuning uses machine-generated instruction-following data to improve LLMs in the language-to-image tasks. It extends instruction tuning to connect vision and language understanding with models like LLaVA (Large Language and Vision Assistant) [76]. LLaVa-1.5 architecture contains a pre-trained visual encoder, a pre-trained LLM to interact with users (Vicuna1.5 with 13 billion parameters [86]), and a vision-language cross-modal connector to align the vision encoder outputs to the language model. The model improved baselines on VQA (Visual question answering) tasks that answer an NL question about visual images. The results also suggest that visual instruction tuning is more critical in improving an LMM's capabilities than pretraining. However, it requires high-quality, targeted visual instruction-tuning data and incurs high computation costs.

Contrastive Learning (CoL) learns meaningful representations from large-scale unlabeled data by maximizing the agreement of the positive pairs formed by the original instance and its corresponding augmented instance while minimizing the agreement with the negative pairs formed by other instances [94]. CoL can also use supervised learning by adding label information based on category discrimination. Some image classification tasks introduce prototypes, which force the embedding of instances closer to their corresponding prototypes in the embedding space while moving them away from other prototypes [95]. Contrastive Language-Vision Pre-training (CLIP) [96] represents a VLM pre-trained on large image-text datasets. The model contains one image and one text encoder. CoL is conducted during pre-training, where a correctly paired image and text is a positive pair. In contrast, incorrectly paired image and text (belonging to distinct image-text pairs) are a negative pair. The prediction is formed by finding the closest text embedding given the image. CLIP is known for its ability to perform zero-shot image classification, which can classify images based on NL descriptions without specific training in those categories [96]. The study from [97] introduced Zero-Shot Continual Learning (ZSCL) to address the CF in continual learning with the VLM through feature distillation and parameters ensemble. Distilling the initial model features on a reference dataset significantly enhances its performance. Parameter ensemble among different training stages effectively mitigates the forgetting issue. The study also proposed a Multi-domain Task Incremental Learning (MTIL) benchmark to evaluate learning tasks from different domains, surpassing state-of-the-art methods. However, the limitation is that an extensive, high-quality reference dataset is needed.

Sequential fine-tuning (SEQ) described in [98] examined fine-tuning PLMs to learn a sequence of 15 downstream tasks of 4 types, including text and intent classification, named entity recognition, and relation extraction. For encoder backbones, the study used BERT (versions with 109 and 335 million parameters). Decoder backbones used GPT2 (124 and 774 million parameters) and GPT-NeoX (19 million to 1,21 billion parameters). The study revealed that the inherent anti-forgetting ability of PLMs comes from the pre-training stage and that classifiers learned in SEQ cause forgetting. Specifically, the relative position changes between the class embeddings in classifiers and the features extracted by PLMs led to decreased performance on old tasks even when PLMs do not forget. Besides, the experiments focused on IL classification tasks.

Class-incremental Learning (CIL) is employed by Vision-Language Models (VLM) to learn general representations from textual information without forgetting previously learned tasks [99]. PROjectiOn Fusion (PROOF) [100] approach alleviates CF in VLM. The model appends linear projections on the pre-trained image/text backbones. The respective projection layer encodes the task-specific information by mapping the projected features. Learning new tasks extends new projections while freezing old ones. The benchmarks indicate PROOF learns new classes while retaining the old ones, with state-of-the-art classification performance. Nevertheless, learning new tasks enlarges the model size and requires more training exemplars.

Processing long input sequences during inference is challenging for LLMs as they perform poorly on inputs whose lengths are larger than those in training sequences. SelfExtend [101] is a method that aims to handle long contexts without fine-tuning. It extends the LLMs' context window by structuring the attention component into a group and a neighbor attention. The group attention manages long-distance token relationships by tracking the dependencies among distant tokens. The neighbor or standard attention tracks dependencies among adjacent or proximate tokens based on a specified range. The attention matrices are merged, and the softmax operation is applied. This way, the method connects the distant positions at inference time to positions seen during training so that LLM can handle longer texts without additional fine-tuning. Extensive evaluations with LLaMA-2 [102] and its successors, Phi-2 [103], Mistral [104], and SOLAR [105], on short-context tasks, synthetic and real-world long-context tasks, and language modeling tasks have consistently demonstrated the effectiveness of SelfExtend. However, this extension in context window length comes at the cost of increased computation, as it requires calculating extra attention across all query-key pairs.

The Scalable Language Model (SLM) [106] adapts the base LLM to new tasks from distinct domains while keeping its performance on the older tasks. SLM integrates the task distribution within the vector space retrieval into the language model for knowledge expansion and management. The Dynamic Task-related Knowledge Retrieval (DTKR) component identifies the most relevant knowledge for each input instance, assuming each task has a distinct data distribution in the vector space. It preserves relevant knowledge by compiling weight increments through low-rank adaptation to decrease computation. The Joint Adaptive Re-parameterization (JARe) then uses these weight increments to obtain adaptive re-parameterization of the pre-trained model, aligning it with downstream tasks based on their distribution. The SLM achieves state-of-the-art performance on different base models, including BERT [107], T5 [108], and LLaMA-2 [109]. However, the method introduces a separate retrieval component, increasing memory and computational costs.

Cross-modal CL [110] integrates multiple modalities via cross-attention, transferring information between different modalities into a common semantic and representation space. Additionally, a dynamic codebook keeps unknown information from incoming modalities as appended dictionary codes. This way, the model can generalize to new tasks and modalities while mitigating the CF of previously learned information. While it outperforms state-of-the-art models on bi-modal generalization tasks, it requires pre-training on multimodal datasets to acquire sufficient representation space.

*Memory-based examples*

Diffusion-based Class Incremental Learning (DiffClass) [111] uses Diffusion Models (DMs) [112], generative models that produce synthetic samples, to improve the classifier performance in CIL tasks. DMs generate images in a stochastic process by progressively denoising them [112]. They incrementally introduce Gaussian noise into the input data (a forward diffusion step). The procedure is gradually reversed to generate synthetic samples, which are used to update the classifier by removing the noise from noised input (a reverse diffusion step). Synthetic samples are similar to the real ones. The DiffClass produces artificial examples using Multi-Distribution Matching (MDM) models to balance the quality of the generated samples and their diversity. Before generating samples, Selective Synthetic Image Augmentation (SSIA) enlarges the training data distribution and enhances the model's plasticity. The method demonstrates state-of-the-art performance in example-free settings in CIFAR100 (60000 images with 100 classes) and ImageNet100 (135,000 images with 100 classes) benchmarks. However, learning new classes prolongs finetuning.

Some approaches [113] introduce and train soft prompts for each new task. However, the accumulation of prompts raises scalability issues by inducing higher training costs.

Instruction-based Continual Learning (InsCL) [114] replays data for each previous task based on their similarity calculated as Wasserstein Distance (WD). The instruction embeddings are produced for pairs of tasks, and the proportions of instructions for each task are calculated as probability distributions. When learning a new task, a replay dataset is formed by sampling examples from previous tasks, where the size of the datasets is estimated as WD between new and previous task training data distributions. The number of previous examples to replay is determined based on their similarity with the task at hand. Moreover, it introduces the Instruction Information Metric (InsInfo) to measure the instructions' quality and diversity and guide the replay process to use high-quality data. Evaluation of 16 distinct tasks (classification, coding, sentiment analysis, QA, comprehension, dialogue and generation among others) demonstrates consistent performance improvements. However, it requires high-quality instructions since fuzzy instructions can affect the task similarity calculation and decrease its performance.

The study [115] examined how pre-trained language models (PLMs) tackle out-of-distribution (OOD) data in CL scenarios. In particular, the model without fine-tuning has poor generalization to OOD data while performing strongly on in-distribution (ID) data. Thus, it must learn from a data stream containing new, unseen examples over time. It fine-tuned two transformer-based architectures (GPT-2 decoder with 110 million parameters and RoBERTa encoder with 135 million parameters) on two downstream tasks (predicting API usage and API call). The datasets with API usage sequences were collected from GitHub as Java programs managing one of the associated APIs (10 million samples as ID to pre-train the models and 147,000 samples as OOD to fine-tune the models continually). Reply-based and regularization-based CL techniques mitigated CF to some extent while maintaining or slightly improving PLMs' performance in downstream tasks. The work builds on a carefully selected ID/OOD scenario and cannot guarantee generalizability on novel, unseen examples.

The study in [116] introduces a CoLeCLIP, a method for Open-Domain CL (ODCL) in VLMs that enables them to learn from a sequence of tasks with non-overlapping classes without forgetting previously learned knowledge. CoLeCLIP uses a class vocabulary and a prompt-based approach to learn task-specific patterns and refine class embeddings while mitigating forgetting through replay-free cross-domain vocabulary learning. It simultaneously learns task prompts and class embeddings in CLIP's text and image encoders [59]. The model outperformed state-of-the-art methods concerning classification accuracy and forgetting on 11 datasets for open-domain CL in both CIL and task-incremental learning (TIL) scenarios. However, real-world ODCL can use very different datasets representing incoming tasks, which may exhibit significant domain gaps and large variations in class correlations and data distributions.

Rehearsal-based methods use past training data, which may be unavailable or nonexistent at inference time. Self-Synthesized Rehearsal (SSR) [117] utilizes the LLM to produce synthetic rehearsal data. It finetunes the base LLM to create synthetic examples through I-CL with few-shot examples. Next, it employs the latest LLM to refine outputs using the synthetic instances. Finally, it selects diverse, high-quality synthetic instances augmented with currently available data for future rehearsals. SSR employs LLaMA-2-7B, LLaMA-2-7B-Chat [109], and Alpaca-7B [118] LLMs using Super-Natural Instructions (SuperNI) [119], a comprehensive instruction tuning benchmark dataset. The results indicate that SSR is more data-efficient and achieves superior or comparable state-of-the-art performance while preventing CF. However, synthetic instances may contain unreliable content from biases in data, degrading performance on unseen tasks.

The study from [120] combined learning rate (LR) re-warm, LR re-decay, and replay techniques to compare the performance with full-data re-training. The study trained the GPT-NeoX [121] decoder-only model and compared the performance by measuring final loss and LLM evaluation benchmarks. The results show that LLMs can be successfully adapted by combining the above elements. The performance matches the re-training baseline with a compute fraction. In particular, re-warming and re-decaying the LR from a large to a small value in pre-training improves adaptation to new tasks. The study also suggests 5% replay as a default value and using more replay with significant distribution shifts. However, the experiments updated the model on only two subsequent tasks.

### B. Meta-Learning

The *meta-learning* (MeL) paradigm goes beyond existing approaches in that a model first learns how to learn (meta-training phase) and then learns new tasks efficiently (meta-testing phase) [122]. It is usually concerned with efficient model adaptation to new tasks quickly, using few examples, and generalization across these tasks [123, 124]. MeL is critical when data is scarce, difficult to obtain, or its distribution constantly changes. Currently, it is analyzed by looking at how NN-based architectures learn reusable representations using a small number of training examples (aka few-shot learning) [122]. This way, tasks share specific structures, enabling models to transfer knowledge across tasks, including image recognition and sentiment analysis [125], machine translation [126], and dialog [127].

NN-based perspective on MeL introduces two components: a *meta-learner* that produces task-specific parameters using a smaller amount of labeled data from a task and a *base learner* that makes predictions on unlabeled test data from the same task [122]. The MeL components and approaches are shown in Figure 3.

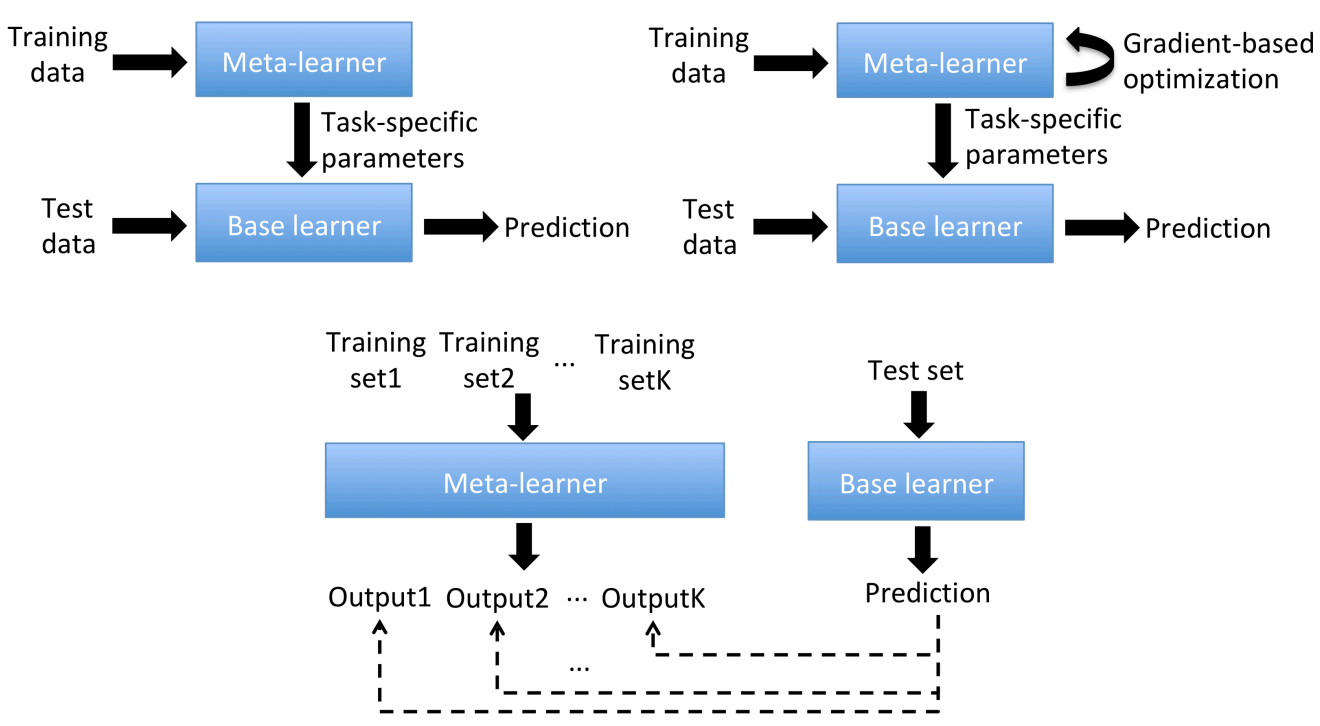

**Fig. 3.** Approaches to Meta-Learning: black-box methods (top left), optimization-based methods (top right), and distance metric learning methods (bottom).

Depending on the implementation of the above components, MeL methods can be categorized as [122, 123]:

1) *Black-box* methods — A black-box neural network takes the entire training dataset and predicts task-specific parameters, which are then used to parameterize the base network and make predictions for test data. The Simple Neural AttentIve Learner (SNAIL) is a typical example [128]. While these methods are versatile regarding tasks and learning problems, the meta-learner architectures are complex with high computational and data requirements.
2) *Optimization-based* methods — the meta-learner is an efficient optimization process (e.g., gradient descent (GD)). It obtains a set of meta-parameters that are easy to learn via GD and fine-tunes them on new tasks. In other words, it learns task-specific parameters from training data and then updates the initial set of meta-parameters by optimizing the performance on the test set of the same task. Model-Agnostic Meta-Learning (MAML) technique works in the described way [129]. The meta-parameters can be inner optimizers, NN architectures, and other network hyperparameters [130, 131]. Nonetheless, backpropagating through multiple gradient adaptation steps is computationally expensive.
3) *Distance metric learning* methods — while the previous two methods use NNs as parametric base models, this method employs parametric meta-learners to produce non-parametric learners (base models). The methods learn an appropriate distance metric (e.g., Nearest Neighbors) for comparing instances in a low-dimensional space, enabling the model to learn an adequate embedding space for classification tasks. Prototypical Networks take the average of a class's examples' embeddings as a prototype and classify unseen examples based on their distances to the prototypes [132]. During the meta-test, each example in the training set is compared with the test instance to determine whether they belong to the same class [133]. These methods may not capture complex relationships between data points.

MeL methods are typically combined with other techniques to extract and represent knowledge from LLMs. We provide an overview of some practical examples, along with their limitations, as follows.

Other approaches to learning across multiple tasks by parameter sharing include *Multitask Learning* (MTL) and *Transfer Learning* (TL). MTL aims to improve task performance by learning them simultaneously [134]. One approach to MTL is *hard parameter* sharing, where the model parameters are split into shared and task-specific parameters [135]. Another approach is *soft parameter* sharing, which measures parameter similarity across task-specific models using regularization penalties [136]. However, choosing the appropriate approach, designing the model architecture, and determining the extent of parameter sharing across tasks depend on the problem being solved [134]. TL fine-tunes a pre-trained model on a new task using less training data [109]. The model leverages representations learned from previous task(s) when solving new tasks. The standard transfer learning approach is fine-tuning, which starts from a pre-trained model whose parameters are then fine-tuned on the training data from the target task using GD or other optimizer functions [137]. However, fine-tuning can destroy initialized features from the pre-training phase [138]. Preventative measures include a lower learning rate for earlier layers, freezing earlier layers, and gradually unfreezing or re-initializing the last layer [139]. Fine-tuning may be ineffective when the target task dataset is insufficiently large [140].

MeL is used in some complex learning scenarios. For example, learning from diverse multimodal task distributions sampled from a task distribution with multiple unknown modes [141]; distributed learning without sharing data across client models as personalized federated learning [142]; learning that adapts to data distribution shifts between the training and testing for more effective domain adaptation-generalization tradeoff [143]; and unsupervised meta-learning using unlabeled data [144].

MeL faces some fundamental challenges. Generalization in which tasks in meta-testing are from a different distribution seen at the meta-training time remains difficult to achieve [122, 123]. Meta-training on multiple data modalities occurs by constructing separate meta-learners focused on tasks from one modality, as different modalities have different dimensionalities [145]. Learning a meta-learner that captures multiple data modalities remains unsolved. Current meta-learning methods have significant computational and memory costs [122, 123]. Thus, deeper theoretical improvements concerning sample complexity, generalization metrics, the design of optimization functions, and proper integration of domain/task-specific knowledge are necessary.

Graphs are crucial in representing real-world objects. Graph Neural Networks (GNNs) learn on graph-like representations with textual node attributes (aka text-attributed graphs) [146]. They represent graph nodes as textual embeddings, exposing limitations in interpreting causal knowledge and understanding semantics [146]. Besides, they are usually retrained to accommodate different tasks on graphs. LLMs are commonly used for graph node classification tasks by enhancing nodes' text attributes [147].

The Graph-oriented Instruction Tuning of Large Language Models for Generic Graph Mining (MuseGraph) [148] extracts compact graph descriptions using graph analysis techniques such as random walks and one-hop neighbors. It textualizes graphs with essential semantic and structural details that LLMs can process under limited input token size. These descriptions are used to produce instructions for different graph mining tasks (e.g., node classification and link prediction). In particular, by designing CoT templates [9] for selected graph tasks and prompting GPT-4 to produce a small set of CoT-based instructions. Finally, the instructions guide graph-aware instruction tuning of LLM for each target graph mining task. It employs LLaMA-7B LLM [109] with LoRA [58] for instruction tuning. The evaluation demonstrates performance improvements in graph mining tasks compared to GNN-based and LLM-based methods. However, it is verified on a limited task coverage and instruction tuning set.

The study from [149] introduces NPHardEval4V, a dynamic benchmark to understand the extent of reasoning capabilities in MLLMs. It is constructed from textual questions in NPHardEval [150] by converting them to images. The NPHardEval benchmark contains 9 different algorithmic problems. The evaluation is conducted on various MLLMs, including GPT-4V [151] Gemini 1.0 Pro [152], CogVLM 17B [153], Qwen-VL-7B [154], and Kosmos2 1.5B [155]. The study experimented with visual, text, and combined prompts. The MLLMs' reasoning performance degrades when increasing question difficulty in individual reasoning tasks. The effects of prompt design combining visual and textual inputs vary across the MLLMs, emphasizing the importance of an appropriate balance when combining visual and textual information.

### *C. Parameter-Efficient Learning*

Parameter-efficient tuning (PET) adjusts a subset of the pre-trained model's parameters by adding new parameters or modifying a smaller subset during the finetuning process [156, 157]. Selective adjustment of a smaller set of parameters reduces computation costs and memory consumption, uses less training data, and is less prone to overfitting. Figure 4 presents PET approaches.

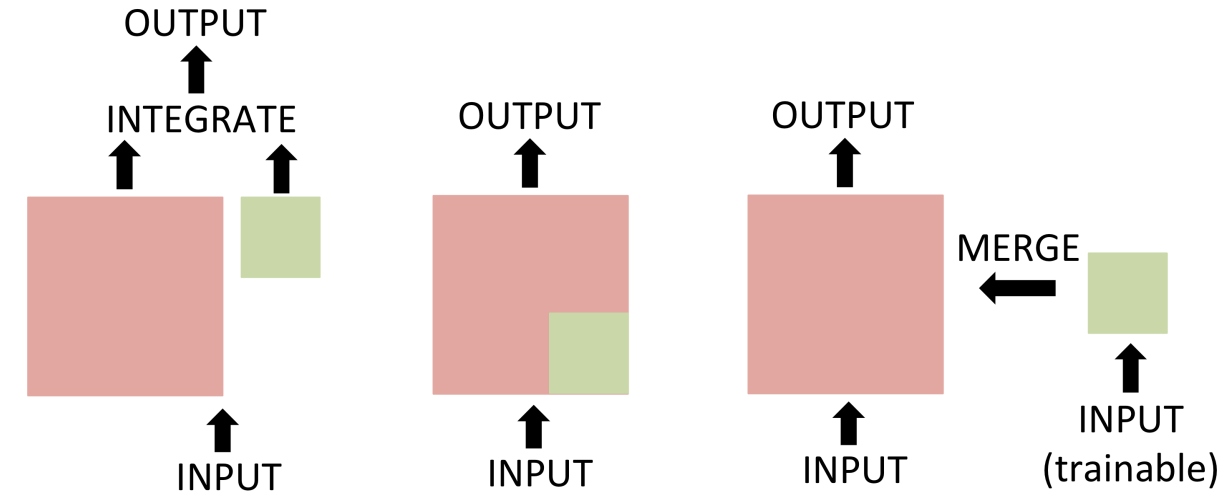

**Fig. 4.** Approaches to Parameter-Efficient Tuning: addition-based (left), specification-based (middle), and reparameterization-based (right). Trainable parts are in green, while the frozen parts are in red. Adapted from [157].

PET for LLMs occurs in the following principal ways [157]:

1) *Addition-based* methods [158] add trainable parameters while keeping the original LLM intact. Specifically, they modify the model structure by injecting trainable parameters or modules. Finetuning updates only the injected parameters, reducing memory and computational demands. Several additive methods exist.
   *Adapter* approaches insert smaller adapter layers within Transformer blocks. *Serial adapters* [159] are positioned after the self-attention and FFN layers. They can reduce the model's parallelism and increase its complexity. *Parallel adapters* introduce adapter layers parallel to Transformer layers [160]. For example, CoDA [160] identifies relevant tokens processed by all layers to maintain overall model performance. The adapter only processes the' unimportant' tokens for efficient inference without compromising overall performance.
   *Soft prompts* are an alternative to adapters in that they prepend adjustable vectors (soft prompts) to input sequences [161]. Similarly, *prefix-tuning* [6] generates learnable vectors as prefixes to all Transformer keys and values. Fine-tuning optimizes the prefix vectors for inference.
2) *Specification-based* methods adjust only a portion of the parameters, such as BitFit, modifying only the model's bias terms or their subset [162]. In particular, they select and finetune a smaller parameter subset for downstream tasks. For example, Diff pruning [163] applies a binary mask by learning a task-specific difference vector, which is added to the pre-trained model parameters that remain fixed during fine-tuning. The mask identifies the finetuned parameters by selective pruning. Structured diff pruning considers dependencies between parameters to modify local groups, whereas unstructured diff pruning treats parameters independently. The former adapts better to the target model and outperforms the latter in terms of accuracy and training efficiency [163].
3) *Reparameterization-based* methods [58] transform a model's architecture by converting LLMs' adaptive parameters into parameter-efficient forms during optimization. Transforming the original model into a functionally equivalent but simplified representation optimizes inference speed. They introduce additional low-rank trainable representations for finetuning combined with the original model. The goal is an effective low-dimensional fine-tuning reparameterization. Low-Rank Adaptation (LoRA) [58] is the widely used technique introducing parameters to model the weight differences and performs better than most mainstream PET methods [156, 164]. Specifically, it constructs two trainable weight matrices. Upon fine-tuning, its weights are integrated with the pre-trained weights to maintain the inference efficiency without extra costs. For example, the PET time on the LLaMA-7B model [109] was approximately one-fourth compared to tuning all parameters [165]. The challenge is choosing an appropriate rank for trainable matrices.
4) *Mixed* methods integrate PET methods and techniques, utilizing their advantages while mitigating drawbacks. UniPELT [166] incorporates prefix-tuning, adapters, and LoRA into LLM blocks, demonstrating consistent improvements in accuracy. S4 [167] explores design spaces to combine and formulate different PET methods as design patterns. LLM-Adapters [168] builds a framework incorporating different PET methods into LLMs. The smaller LLMs utilizing PET and appropriate finetuning data show competitive performance compared to training a larger parameter set.

The PET methods, along with their affected model components and elements, as well as their trade-offs, are summarized in Table 3.

TABLE 3. PET METHODS COMPARATIVE SUMMARY.

| Method | Component | Element | Trade-offs |
|---|---|---|---|
| Addition-based | Adapters<br>Soft prompts | Parameter layers<br>Learnable vectors | Updates injected parameters<br>Increases model size |
| Specification-based | (Un)Structured Masking | Parameter subsets | Updates portion of parameters<br>Intricate parameter selection |
| Reparameterization-based | Low-rank Representations | Parameter-efficient forms | Efficient parameter representation<br>Non-trivial representation choice |

PET methods are commonly part of larger CL solutions. We outline some typical examples as follows.

Continual Parameter-Efficient Tuning (ConPET) is an extension for continual LLM learning [169]. Static ConPET combines data replay with PET. Dynamic ConPET comprises lightweight, task-specific PET modules as experts tuned without changing the original LLM parameters. The results demonstrate a significant reduction in tuning costs. However, this remains to be verified using more diverse continual learning scenarios and task split strategies among experts.

The study from [170] introduces a PET framework for vision-language models for CL to alleviate CF. It expands a pre-trained CLIP model [59] by adding LoRAs [58] as sparse experts in all image and textual encoders. The Distribution Discriminative Auto-Selector (DDAS) routes inputs to the expert adapters or the pre-trained CLIP. Task-specific routers activate corresponding experts responsible for particular tasks. The framework outperforms state-of-the-art solutions on classification tasks and reduces parameter training costs. However, the number of experts increases with the number of tasks, and it does not improve the backbone CLIP.

The SSF (Scale and Shift the deep Features) method [171] scales and shifts the pre-trained model's features to learn new tasks, if upstream and downstream datasets have different data distributions. Scaled and shifted parameters fall in discriminative and unified parameter space, as variance and mean, to modulate the trainable features for downstream tasks. The method is evaluated on different pre-trained MLLM backbones, including ViT-B/16 [59] and Swin Transformer [65]. It outperforms other PET methods. However, obtaining tunable parameters by scaling and shifting original parameters needs verification in domains whose training data significantly vary from the ones used in pretraining.

Post-deployment LLM updates to deal with errors and capture changing data is known as Model Editing (ME). The MELO [69] represents a plug-in ME solution integrating LoRA modules indexed in an internal VD. It extends backbone LLM capabilities by activating specific LoRAs based on their VD indices. LoRA blocks are activated by searching inputs in the VD during inference. LoRA blocks from different layers but shared indices are active, and training updates VD clusters for prospective LoRA searches during inference. The MELO employs multiple LLM backbones for editing, including BERT, T5-Small, and T5-Large. Experiments on tasks of classifying documents, QA, and correcting hallucinations indicate state-of-the-art performance while requiring the least trainable parameters and computational cost. However, its size increases with the number of learning tasks and introduces a VD as an additional component to the backbone models.

While existing PET methods focus on updating a small number of LLM parameters, the study from [172] develops a family of Representation Finetuning (ReFT) methods updating hidden LLM representations. They learn task-specific functions, manipulating a smaller fraction of hidden representations to solve new tasks during inference. ReFT instance called Low-rank Linear Subspace ReFT (LoReFT) finds internal representations from a low-rank projection matrix using the Distributed Alignment Search (DAS) [173]. DAS finds an optimal alignment between an interpretable high-level causal model and a low-level DL system it simplifies using GDs. LoReFT is evaluated on LLMs of different sizes, from RoBERTa 125M [174] to LLaMA 13B [109], and benchmarked against existing PET methods. It uses ten times fewer parameters than LoRA, while showing comparable performance.

The Shared Attention Framework for Parameter-efficient conTinual learning (SAPT) [175] introduces the Shared Attentive Learning & Selection module (SALS). Specifically, the SALS pre-processes training samples to find the optimal PET block combinations for completing the current task. It does so by obtaining an attention weight via instance-level shared attention operation. Inputs follow the same shared attention operation to select the appropriate PET blocks. The Attentive Reflection Module (ARM) assists SALS in recalling the shared attention operation of samples of earlier tasks that should be performed initially through generative replay with synthesized samples to prevent forgetting. The SAPT has been evaluated on the SuperNI CL NLP benchmark [119], including dialogue generation, information extraction, QA, summarization and sentiment analysis tasks, and Long Sequence [93], a CL benchmark with 15 classification tasks. It achieved superior performance compared to other PET methods with different model sizes (from 770M to 13B) and architectures, including the T5 [108] and the LLaMA-2 [109]. However, increasing the learning sequence (number of tasks) requires additional PET blocks.

### D. Mixture of Experts

MoE architectures extend LLM capabilities by combining multiple layers or components as experts. The router mechanism determines which experts are activated for a particular task. Figure 5 shows typical MoE architecture.

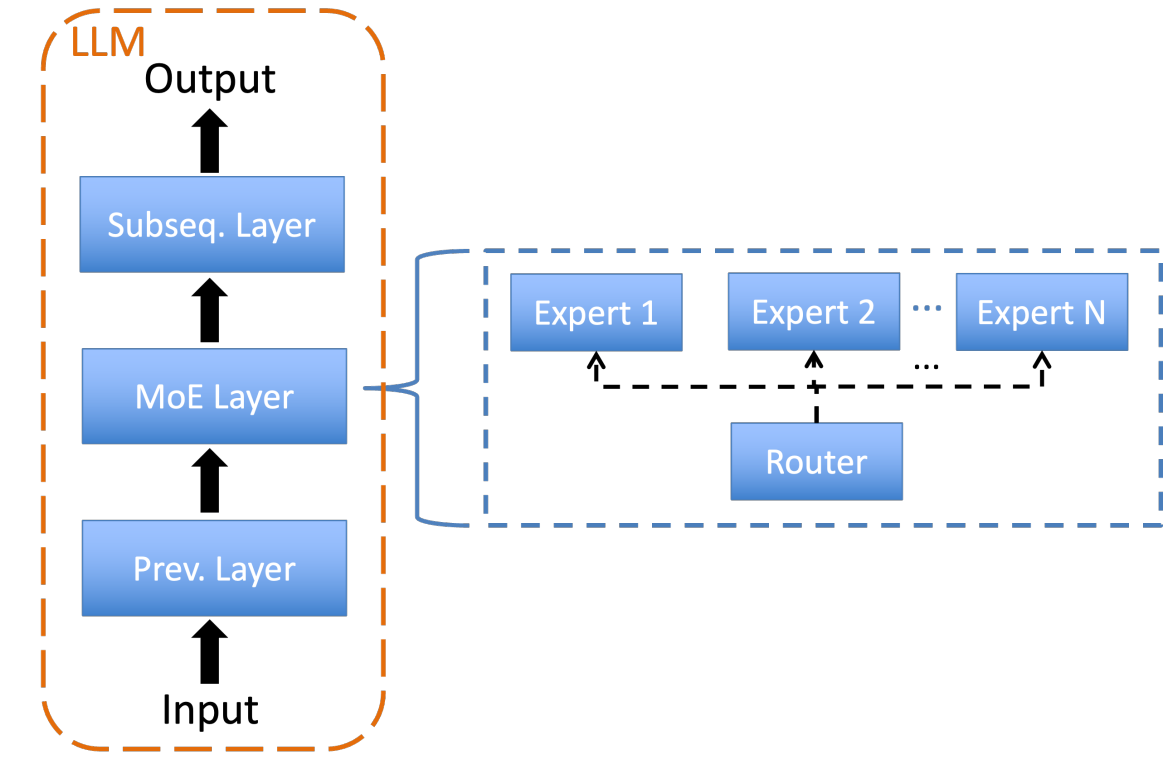

**Fig. 5.** Mixture of Experts in LLMs.

Experts are mainly model layers' components (such as FFNs and attention heads) [176]. Specific inference tasks are delegated to one or more expert FNNs in a respective model layer through a gating mechanism (i.e., router) for each input. Specifically, Dense MoE activates all experts from the layer, whereas Sparse MoE selects a subset of experts (e.g., top-k) [176].

The MoE approaches, along with their respective model elements and trade-offs, are outlined in Table 4.

TABLE 4. MOE APPROACHES COMPARATIVE SUMMARY.

| Expert | Element | Trade-offs |
|---|---|---|
| LLM layer component (FFN, Attention) | Parameters<br>Activation functions<br>Attention heads | Straightforward task delegation<br>More experts affect routing efficiency<br>New tasks increase model size and computation |
| PET (FFN and Attention low-rank decomposition) | Efficient representations (parameter, activation, attention) | Smaller size<br>Accuracy depends on expert selection quality (i.e., routing)<br>Routing increases latency<br>New tasks increase model complexity |

MoEs are employed in various tasks and combined with previously described CL strategies, as illustrated by the following examples and their trade-offs.

Domain expert mixture (DEMix) [177] is a collection of domain-specialized expert FFNs. The experts are added, removed, or mixed after pretraining. However, introducing additional layers modifies existing structures in Transformer-based architectures [1] (e.g., FFN layer). Besides, it keeps all experts in memory and can face scalability issues.

The study from [178] employed instruction tuning on MoE models consisting of FFN experts. While instruction tuning on downstream tasks improves the model capabilities, it requires high-quality instruction examples and larger models.

The research in [179] combined MoE with other methods, such as PET and instruction tuning on 'lightweight' experts (updating less than 1% of an 11B parameter model). However, the parameter percentage (or expert size) varies with the pre-trained model size and the number of experts involved.

Another study trained domain-specific LoRA modules as Mixture-of-LoRAs (MoA) [180]. The modules are aligned using a predefined routing strategy. However, expert training can require labeled data at scale to specialize in a specific domain and the adaptation of the routing algorithm.

Similarly, Mixture of LoRA Experts (MoLE) [181] models experts as layers containing trained LoRAs. Moreover, it learns a gating function within each expert (layer) for efficient and flexible selection of multiple LoRAs. Although it outperforms existing LoRAs compositions, adding new LoRAs (up to 128) decreases composition performance.

The Mixture of Prompt Experts (MoPE) [182] is a conditional multimodal prompt tuning method. MoPE conducts multimodal input fusion by introducing multiple prompt experts rather than extending individual prompt lengths. It learns multiple fixed-length prompts as experts and an expert router per layer. The router selects the appropriate prompt for each input instance based on its modality-specific embeddings. Its results are comparable to fine-tuning, requiring less parameters. However, increasing the number of tasks increases the number of experts, and different routing strategies affect expert learning.

Finally, the study in [183] extended Mistral 7B LLM such that each layer comprises eight FFN blocks as experts. Two experts process the current state and combine their outputs for every token at each layer. Each token can access 47 billion parameters but only uses 13 billion active parameters during inference. Although it demonstrates performance and efficiency gains concerning other LLMs, its capability to learn new tasks depends on the model size (number of parameters).

## III. PRINCIPAL FINDINGS

### A. Continuous Learning

CL learns incremental tasks with dynamic data distributions in various domains, including object detection and semantic segmentation (computer vision), NLP tasks, conditional generation (generative models), and autonomous driving (perception, motion planning, motion control, and user-vehicle interaction) [2].

One of the key challenges in CL is to achieve an appropriate balance between stability and plasticity while ensuring adequate generalizability intra-task (from training to test data) and inter-task (accommodating data distribution changes) [22, 157]. Stability concerns LLM's memory as the difference between the task's current and maximum past performance (aka forgetting measure) and how learning a new task affects learned tasks (aka backward transfer). Plasticity is an LLM learning capacity, defined as the extent to which the model cannot learn new tasks or the joint influence of learned tasks on learning the current task.

The CL techniques working in embedding space (e.g., soft prompting, adapters and prefix-tuning) are more expressive and effective than prompting which operates in the discrete token space. However, the existing work focuses on learning plasticity and inter-task generalizability [20, 22, 23, 25].

The challenges of continuous learning of LLMs include:

1) *Multimodal Learning* — LLMs are pre-trained with enormous datasets and can pick up very complex and unexpected behaviors (e.g., hallucinations). Real-world data are not restricted to a specific modality but a random combination of multiple ones. Integrating and merging various modalities into a coherent CL workflow requires efficient modalities encoding, multi-modal databases, assigning weights during fusions, and comprehensive benchmarks, which inevitably leads to increased computational costs without general performance guarantees [184].
2) *Data Scaling and Quality* — CL methods require task-specific datasets using sources similar to old tasks [185]. The size and quality should adapt to obtain an appropriate trade-off between the efficiency and the effectiveness of learning incremental tasks. Inherent biases in training data can manifest in harmful or incorrect model decisions [186]. Besides, the transferability of the task- and domain-specific training data and their compositionality with other tasks and domains without extra inference cost and scalability issues in large-scale training remain unsolved. Finally, effective and efficient CL methods may differ for smaller and larger models. Exploring scaling strategies that adapt and remain effective as the model size varies is crucial due to the ever-evolving landscape of LLM architectures.
3) *Sustainability* — The environmental impact of training and deploying LLMs is still at stake due to enormous resources

required. Future CL models and pipelines should be more energy-efficient and operate in low-resource environments to reduce environmental footprint and increase their accessibility [187].

4) *Trustworthiness* — LLMs are susceptible to adversarial attacks. Modifying inputs can produce unexpected and harmful outputs [188]. CL system design should include encryption protocols for personal data and intermediate training and inference results without compromising performance. Improving CL models' robustness against such attacks is important for safety-critical applications.
5) *Few-Shot Generalization* — LLMs struggle to learn tasks from few examples or domain-specific knowledge [189]. CL should improve its performance with limited training data. These data should be carefully selected as their distribution shift can degrade performance compared to previous data or cause poor model adaptation.
6) *Multilingual Learning* — Making CL more accessible and effective in underrepresented languages with limited training data [190].
7) *Responsiveness* — The inference in CL should generate predictions in real time for applications such as chatbots or recommenders, where low-latency responses are critical [191]. LLM inference efficiency is affected by the model size, increased computational costs of longer inputs, the number of requests, and dependencies on external knowledge sources. Rethinking what constitutes efficiency is necessary as additional parameters and representations are computed and stored. Model compression techniques [192] (e.g., pruning and quantization) focus on the LLM's parameters representation, not their architectural elements.
8) *Context Size* — CL uses a limited amount of input tokens to generate subsequent words due to LLMs' constrained context window [193]. This can cause difficulties processing lengthy sequences concerning coherence and relevance since the model can neglect or lose track of the relevant information.
9) *Up-to-date Knowledge* — LLMs learn from data snapshots, and their knowledge is restricted to what is available on a particular date. Consequently, CL needs access to the most recent information. At the same time, updating LLMs with up-to-date knowledge may outdate their internal knowledge [194]. Besides, managing conflicting knowledge from one or more sources to choose the correct one is non-trivial for LLMs and CL. For example, different datasets can contain contradictory information about the same fact or carry conflicting biases and inaccuracies. Currently, external resolution strategies must assess the data's reliability.
10) *Unified benchmarking* — Various benchmark tools and datasets make it difficult to reliably compare the performance of different methods and algorithms [195]. There are no guarantees that benchmark data were not used for LLM training directly or indirectly. Moreover, the performance against research benchmarks does not necessarily imply usefulness in a real-world setting. These may be small for industry standards, and data can be very different [195]. Open-ended evaluation is the biggest bottleneck to AI adoption, and current efforts focus on solutions that can be evaluated [195].

A recent study [196] discovered that popular context-based parameter-efficient fine-tuning methods, including prompting, in-context learning, prompt tuning, and prefix-tuning, are potentially less effective than full fine-tuning, even with the same number of learnable parameters. In particular, they have structural limitations when learning new attention patterns since they cannot change the attention pattern over the content and can only drive the outputs of an attention layer in a fixed direction. Therefore, despite their success in learning tasks on the fly, they may be unable to learn novel tasks that require new attention patterns.

Table 5. compares and exemplifies previously analyzed IL approaches and their methods against the given criteria. Task-incremental support denotes the capability of learning a sequence of tasks, each with a known identity that is leveraged during training and inference. Parameter update volume is the fraction of the parameters that are updated. Real-time adaptability refers to the ability to adapt promptly to unseen data or tasks. Historical data dependency indicates the extent to which methods utilize past training examples or representations when learning new tasks.

TABLE 5. IL APPROACHES COMPARATIVE SUMMARY.

<table>
<tr><th>Approach</th><th>Method</th><th>Task-incremental support</th><th>Parameter update volume</th><th>Historical data dependency</th><th>Real-time adaptability</th><th>Example</th></tr>
<tr><td rowspan="3">Continual Learning</td><td>Consolidation-based</td><td>Medium</td><td rowspan="2">< 5% – 10%</td><td>Low</td><td>Low</td><td>PLoRA [55], FSCIL [60], IVLOD [64]</td></tr>
<tr><td>Dynamic-architecture-based</td><td>High</td><td>Medium</td><td>Medium</td><td>MiniGPT-4 [84], SEQ [98], SLM [106]</td></tr>
<tr><td>Memory-based</td><td>High</td><td>~ 10% – 100%</td><td>High</td><td>High</td><td>DiffClass [111], InsCL [114], SSR [117]</td></tr>
<tr><td rowspan="3">Meta-Learning</td><td>Black-box</td><td>Low</td><td>< 5%</td><td>Low</td><td>High</td><td>SNAIL [128]</td></tr>
<tr><td>Optimization-based</td><td>High</td><td>~ 100%</td><td>High</td><td>Medium</td><td>MAML [129]</td></tr>
<tr><td>Distance metric learning</td><td>Medium – High</td><td>< 10%</td><td>Medium</td><td>Medium – High</td><td>Prototypical Networks [132]</td></tr>
<tr><td rowspan="3">Parameter-Efficient Learning</td><td>Addition-based</td><td>Medium</td><td>~ 0,1% – 3%</td><td>Medium – High</td><td>Medium – High</td><td>CoDA [160], Soft prompts [161]</td></tr>
<tr><td>Specification-based</td><td>Medium</td><td rowspan="2">~ 0,1% – 1%</td><td>Medium</td><td>Medium</td><td>BitFit [162], Diff pruning [163]</td></tr>
<tr><td>Reparameterization-based</td><td>High</td><td>Medium</td><td>Medium – High</td><td>LoRA [58], LoReFT [172]</td></tr>
<tr><td rowspan="2">Mixture of Experts</td><td>LLM layer component</td><td rowspan="2">High</td><td rowspan="2">~ 0,1% – 1%</td><td rowspan="2">Medium – High</td><td rowspan="2">High</td><td>DEMix [177], Mixtral of Experts [183]</td></tr>
<tr><td>Low-rank component</td><td>MoA [180], MoLE [181]</td></tr>
</table>

### B. LLMs versus Humans

LLMs engage in an iterative communication process, generating streams of tokens. They rely on underlying probabilistic models to produce token arrays in response to token arrays. When presented with an input token string, they utilize previously calculated weights to iteratively generate and return a token stream that aligns with the input based on a certain probability threshold. This is constrained by the maximum number of tokens LLMs accept at once (context size) to generate the output. External data can be incorporated as vector embeddings and query vector databases (VDs) to augment the trained model during inference (e.g., RAG).

Mainstream LLM research focuses on improving capabilities within existing knowledge boundaries. Understanding the boundaries of their knowledge, referred to as self-knowledge, is essential. Evaluating 20 available LLMs' self-knowledge [197] by measuring their ability to identify questions they do not know or cannot answer revealed a considerable gap between their capabilities and the human ability to recognize the limitations of their knowledge. While humans can quickly acquire new skills by leveraging prior experience (or mental models) as generalists, LLMs are still specialists performing well specific tasks.

Hyperparameters (e.g., adapter size, LoRA's rank, placement of added components, optimizer choice, and number of fine-tuned layers) influence the CL performance. Performance ultimately depends on the manual tuning of hyperparameters, which requires considerable effort. Automated optimization of hyperparameters is still an issue for LLMs, lacking simpler solutions with better performance-efficiency tradeoffs [198].

Current ME approaches aim to make the existing model architectures more flexible to provide cost-efficient targeted updates without retraining the entire model. However, changing one model component can affect other components. Ensuring partial modifications do not reduce general performance or introduce algorithmic- and data-related biases is challenging. VDs are helpful as a cost-efficient information retrieval solution within LLMs' frameworks [15]. However, they are still not optimized for retrieving the information with the intended meaning, such as incremental query adjustment and free-text search using keywords in established retrieval paradigms (e.g., relational and graph databases).

LLMs can store massive amounts of real-world knowledge through intensive training. However, even with the existing IL approaches, they still cannot update their knowledge fast enough to keep pace with constant external, real-world knowledge change and emergence. Accordingly, the IL approaches described and existing solutions [199, 200] still need improvement concerning computational demands and effectiveness. Moreover, KGs mainly rely on textual graph structures to represent knowledge, whereas real-world knowledge inherently comprises different modalities. Effective graph representations for multimodal knowledge that represent and align different modalities into coherent entities remain an issue for KGs [201]. MLLMs encode and align distinct modalities in a vectorial form. This gap between multimodal KG and LLM representations hampers their synergy.

### C. LLMs and Metacognition

LLMs can answer questions and generate useful content on a wide range of topics.

Prompting techniques guide them to generate outputs in response to specific types of inputs. However, they expose shallow reasoning and are prone to errors as their complexity increases. They fail to address complex knowledge because they do not actually know how they learned to produce outputs. LLMs generate statistical responses emerging from numerical weights calculated at training time.

Adapting LLMs to evolving knowledge is challenging LLMs. This requires comprehensive and responsible guidance to incorporate emerging knowledge in a timely manner while conforming to ethical norms to ensure safe usage.

The language modeling paradigm is based on the next token prediction, but its underlying structures and processes remain opaque. Thus, it is unclear whether current KE methods constitute meta-learning stemming from variable probability distributions of training data. Therefore, achieving meaningful and intentional KE with LLM is questionable. For example, a failure in LLM's logical deduction is known as the reversal curse [202], where if a model is trained on a pattern 'A is B,' it will not necessarily generalize in the reverse direction 'B is A.' KE in LLMs may be more effective in task-specific scenarios, where the consequences of model changes can be anticipated.

Knowledge injection usually leads to overfitting LLMs as they struggle to obtain recent factual knowledge via unsupervised fine-tuning [203]. A key challenge is the timely injection of relevant and updated knowledge in LLMs to develop user trust. Maintaining knowledge persistency in LLMs is another challenge, as their current knowledge is represented by generally nonpersistent parametric memory and derived internal representations influenced by inherent CL-CF tradeoff.

LLMs can sometimes solve reasoning problems, and they do so because of the token prediction process. In other words, reasoning problems are translated into pattern completion problems. Building trustworthy reasoning systems using LLMs means integrating them with a logic tool that implements causal mechanisms [16, 17]. Otherwise, we should be able to reverse-engineer their responses to understand an internal process that leads to the reasoning-driven response [204]. Relatedly, the amount of LLMs generalizability corresponds to the extent of token prediction, which ultimately depends on the number of trained parameters and dataset size.

## IV. Conclusion

Significant advancements have been made in developing specialized LLM-based systems to solve specific problems. However, the goals of generality and adaptability across multiple tasks remain an open problem. IL aims to learn new knowledge over time, transfer knowledge across tasks, and efficiently adapt to novel domains.

The review investigated various IL approaches in LLMs that have demonstrated promising results. Nonetheless, open problems and challenges remain, calling for further investigation. Periodic batch updates of LLMs still fail to achieve the IL in real-time. Foundational advancements in IL will enable more versatile algorithms that learn multiple tasks, generalize across domains, and continuously accumulate knowledge.

The broader contribution of the review lies in framing IL beyond merely as a technical phenomenon of LLM tuning, but as a system-level challenge that spans architectural, optimization, and knowledge representation concerns. Accordingly, we see our work as a step forward in encouraging research in this direction. By analyzing the current state of IL in LLMs and highlighting the challenges and opportunities, we aim to inspire researchers and practitioners to explore synergies between and within AI and non-AI research fields for the future progress of IL.

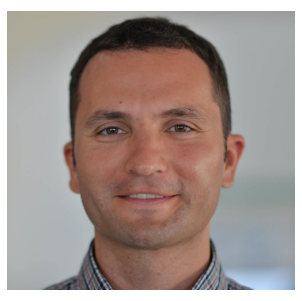

**Mladjan Jovanovic** is an associate professor of Computer Science at Singidunum University, Belgrade, Serbia. He is a visiting scientist at the University of Trento, Italy, and Vienna University of Technology, Austria. He has over 19 years of research and industrial experience in designing, developing, and testing intelligent interactive computing systems. His primary interests include language modeling and knowledge technologies.

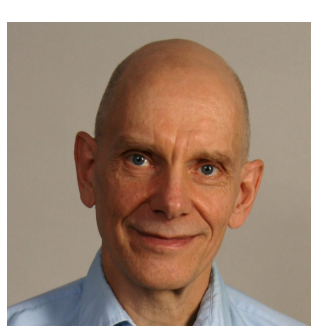

**Peter Voss** is the CEO and Chief Scientist at AIGO.ai, Austin, Texas, USA. He is one of the founders of the term AGI (Artificial General Intelligence) in 2002. Since then, he has been steadily developing, commercializing, and improving the technology geared towards human-level AI. His main interests are cognitive architectures based on a deep understanding of the requirements of human intelligence.